\documentclass[letterpaper, 10 pt, conference]{ieeeconf}  % Comment this line out if you need a4paper

\IEEEoverridecommandlockouts                              % This command is only needed if 
\usepackage{graphicx} % for pdf, bitmapped graphics files 图片

\usepackage{amsmath} % assumes amsmath package installed 数学公式
\usepackage{amssymb}  % assumes amsmath package installed
\usepackage{cite} % 引用： 
\usepackage[colorlinks,linkcolor=red]{hyperref} % 超链接
\usepackage{color}
\usepackage{booktabs}

\title{\LARGE \bf
Real-Time LiDAR Point Cloud Compression and Transmission for Resource-constrained Robots
}

\author{Yuhao Cao, Yu Wang and Haoyao Chen, $Senior\ Member$, $IEEE$% <-this % stops a space
\thanks{The authors are with the School of Mechanical Engineering and Automation, Harbin Institute of Technology Shenzhen, P.R. China. {\tt\small 19B953030@stu.hit.edu.cn.}}%
\thanks{This work was supported in part by the National Natural Science Foundation of China under Grant U21A20119, as well as the shenzhen Science and Innovation Committee under Grant RCJC20231211090050082 and Grant JCYJ20241202123714019. (Corresponding author: Yu Wang)}
}
\begin{document}

\maketitle
\thispagestyle{empty}
\pagestyle{empty}

%%%%%%%%%%%%%%%%%%%%%%%%%%%%%%%%%%%%%%%%%%%%%%%%%%%%%%%%%%%%%%%%%%%%%%%%%%%%%%%%
\begin{abstract}
LiDARs are widely used in autonomous robots due to 
their ability to provide accurate environment structural information. 
However, the large size of point clouds poses challenges in terms of data storage and 
 transmission.
In this paper, we propose a novel point cloud compression and transmission framework for 
resource-constrained robotic applications, called RCPCC.
We iteratively fit the surface of point clouds with a similar range value 
and eliminate redundancy through their spatial relationships.
Then, we use Shape-adaptive DCT (SA-DCT) to transform the unfit points 
and reduce the data volume by quantizing the transformed coefficients.
We design an adaptive bitrate control strategy based on QoE 
as the optimization goal to control the quality of the transmitted point cloud.
Experiments show that
our framework achieves compression rates of 40$\times$ to 80$\times$ while
maintaining high accuracy for downstream applications. our
method significantly outperforms other baselines in
terms of accuracy when the compression rate exceeds 70$\times$. Furthermore, in situations of reduced communication bandwidth,
our adaptive bitrate control strategy demonstrates significant
QoE improvements. The code will be available at \underline{https://github.com/HITSZ-NRSL/RCPCC.git}.
\end{abstract}

%%%%%%%%%%%%%%%%%%%%%%%%%%%%%%%%%%%%%%%%%%%%%%%%%%%%%%%%%%%%%%%%%%%%%%%%%%%%%%%%
\section{INTRODUCTION}
\begin{figure}
    \centering
    \includegraphics{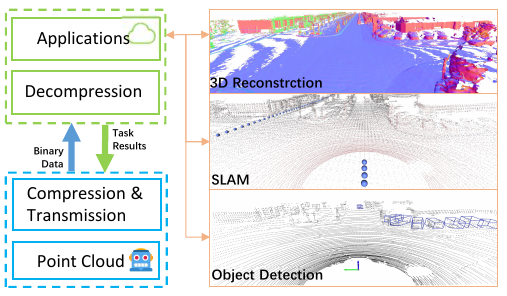}
    \caption{A cloud service solution diagram for resource-constrained robots (left). And the downstream tasks results using the compressed point cloud (right).}
    \label{fig:cover}
    
\end{figure}

Field robots are typically equipped with LiDAR to 
perceive the surrounding environment 
and generate high-precision structural information.
This information is utilized for spatial perception tasks such as object detection and 3D 
reconstruction. 
However, these tasks are often 
constrained by the limited computational resources of 
field robots, especially in scenarios that require 
real-time processing.
To address this issue, as shown in Fig.~\ref{fig:cover}, the combination of edge computing
 and cloud computing provides a feasible solution, 
 by transmitting point cloud data to remote servers or 
 the clouds to perform computationally intensive tasks.
However, the vast amount of data 
generated by LiDAR, often reaching millions of 
points per second, presents a technical challenge 
for real-time transmission.
Moreover, communication links in field environments are 
often unstable, and  bandwidth is limited, further 
exacerbating the difficulty of data transmission.
Therefore, how to achieve efficient point cloud compression 
and transmission under limited computing and 
communication resources remains a critical 
challenge.

Previous research has primarily focused on compression
 rate and point-to-point accuracy~\cite{c14,lasserre2019using,tu2016compressing}, 
 but many of these methods are difficult to apply 
 to resource-constrained robots due to their lack of 
 real-time performance or reliance on heavy 
 computation (e.g., requiring GPUs). 
 Our work focuses on improving compression rate, 
 compression speed, and application-level accuracy 
 (e.g., for odometry and object detection). 
 To ensure low end-to-end latency for applications, 
 we introduce online control of compression quality, 
 ensuring low latency and stability throughout 
 the point cloud transmission pipeline.

We propose a real-time LiDAR point cloud compression 
and transmission framework for resource-constrained 
robots, named RCPCC, 
which achieves a high compression rate, 
maintains high application-level accuracy, 
and operates at a speed ($>$10 Hz) that exceeds 
the LiDAR point cloud generation rate, 
enabling computationally constrained robots to offload 
computation-intensive tasks to the cloud.
To address bandwidth limitations and fluctuations during transmission, 
we propose a QoE-based adaptive bitrate control strategy 
that adjusts the transmission quality based on 
the current and historical buffer queue lengths, 
ensuring optimal QoE and guaranteeing real-time and 
stable point cloud transmission.

Our compression method uses range images as the basic representation of point clouds. Range images utilize the inherent physical properties of LiDAR,
projecting 3D point clouds into 2D images,
enabling high computational efficiency for subsequent 
processing. We iteratively fit surface models to the point 
clouds,
fully leveraging the spatial characteristics of range images 
for point cloud encoding.
For points not fitted by the surface model, we apply Shape-adaptive Discrete Cosine Transform (SA-DCT)~\cite{karam1991adaptation} to the unfit points, 
avoiding zero-value noise artifacts\cite{chenghui50} and achieving further compression.

% We evaluate our method on the KITTI dataset, 
% and experimental results demonstrate that our method outperforms 
% all baselines in downstream tasks when the compression rate exceeds 40$\times$.
%  Our adaptive bitrate control strategy significantly improves QoE and 
%  reduces latency under varying communication quality.
 \par
The main contributions of this paper are three-fold: 
 \begin{itemize}
    \item A novel LiDAR point cloud compression and transmission framework, named RCPCC, is proposed to handles bandwidth fluctuations and enhances the real-time performance and stability of point cloud transmission. 
    \item An adaptive bitrate control strategy based on QoE is proposed to handles bandwidth fluctuations and improves the real-time performance and stability of point cloud transmission. 
    \item Extensive experiments demonstrate that, compared to state-of-the-art methods, RCPCC achieves a competitive compression rate and high accuracy while significantly reducing transmission latency.
 \end{itemize}
\section{RELATED WORKS}
In this section, we reviews the research status of point cloud compression and transmission from the perspective of point cloud encoding and adaptive bitrate streaming.
\subsection{Unstructured Point Cloud Encoding}
Over the past few decades, significant progress 
has been made in point cloud compression. 
However, not all point cloud compression techniques 
are suitable for LiDAR point clouds. 
LiDAR point clouds have unique characteristics, 
such as sparsity, large coverage areas, 
and uneven density, making many compression 
methods designed for 3D object point clouds less 
effective. The most common and widely used methods 
for unstructured point cloud compression are based 
on spatial subdivision trees, such as octrees. 
These methods \cite{c1}\cite{c2}\cite{c3}\cite{c4} 
first use octrees as the foundational data 
structure to structure the point clouds, 
then apply various transformations and 
encoding schemes to reduce spatial and 
informational redundancy. 
In addition to octrees, 
clustering-based and segmentation-based
 methods \cite{c6}\cite{c5} have also 
 demonstrated effectiveness in lossy geometric 
 compression. 
To better capture geometric patterns, 
neural network-based methods \cite{c7}\cite{c8} learn 
the latent spatial structure and geometric information 
within the data, achieving superior lossy and lossless 
compression. However, these methods typically 
lack real-time capabilities and computational 
efficiency. 
Despite their effectiveness in many scenarios, 
the  intrinsic physical properties of LiDAR sensors make LiDAR 
point clouds structured, 
and unstructured methods fail to 
take full advantage of this.
\subsection{Structured Point Cloud Encoding}
In structured point cloud compression, 
many works \cite{flicr} project point clouds into 2D images 
for compression. 
Houshiar et al. \cite{c14} used panoramic cylindrical 
projections to convert point clouds into 2D images 
for lossless and JPEG lossy compression. 
Some works, such as the MPEG V-PCC framework, 
establish local reference frames on point cloud 
surfaces and generate orthogonal projection images 
from multiple views \cite{c10}\cite{c11}. 
These projection images are further compressed 
using existing image or video compression techniques.
However, image or video compression methods 
like JPEG or H.255 are typically designed 
for rectangular data with values between 0 and 255, 
without invalid pixels. 
Direct application to range images can introduce 
significant quantization errors 
and noise \cite{huicheng},
 leading to a reduction in downstream 
 application accuracy.

\subsection{Adaptive Bitrate Streaming}
Common adaptive bitrate (ABR) schemes are crucial 
for video streaming \cite{survey}.They allow 
video streams to dynamically adjust video quality 
according to user network conditions, thereby 
avoiding video stuttering caused by network 
fluctuations. 
ABR methods aim to optimize Quality of Experience (QoE) 
by making optimal decisions. 
ABR methods can be broadly categorized 
into rate-based \cite{festive}, 
buffer-based \cite{bola}, hybrid \cite{mpc}, and 
RL-based \cite{rl} approaches.
Some researchers have explored ABR 
for point cloud data transmission 
\cite{c16}\cite{c17}. 
Unfortunately, these methods were designed for 
dense 3D object point clouds and cannot be directly 
applied to LiDAR point clouds.
\section{METHODOLOGY}
\subsection{System Overview}
\begin{figure*}[!ht]
    \centering
    \includegraphics[width=1\linewidth]{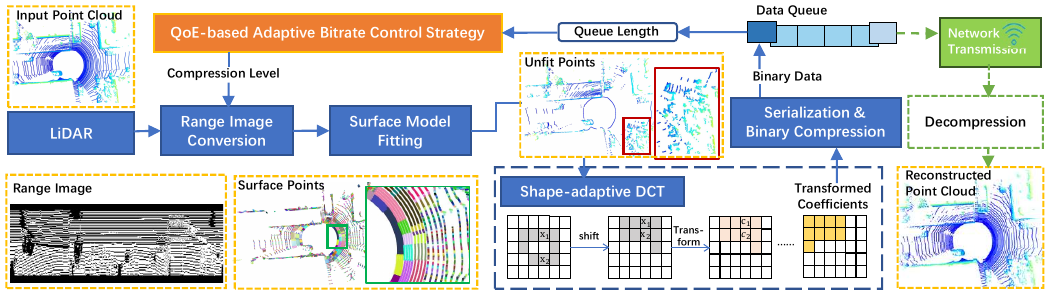}  % 这是一个空的矩形框，大小为 10cm x 6cm
    \caption{Overview of the proposed \textit{RCPCC} framework. 
    The input point cloud is first converted into 
    a range image to accelerate the compression 
    process. We use surface model fitting to 
    eliminate spatial redundancy in the point cloud. 
    Unfit points are transformed from the time domain
     to the frequency domain using SA-DCT, 
     and the transformed results are quantized. 
     Finally, all data required for decompression is 
     serialized and fed into the binary entropy encoder.}
    \label{fig:banner}
\end{figure*}

The proposed real-time LiDAR point cloud compression and transmission framework for resource-constrained scenarios, RCPCC, is illustrated 
in Fig.~\ref{fig:banner}. Point clouds are first projected 
into range images using spherical coordinates. 
Subsequently, the range image is divided into macroblocks, 
and surface model fitting is performed on each macroblock. 
Points in fitted macroblocks are encoded and parameterized,
 while the fitted points are removed from the range image. 
 For the unfit points, SA-DCT is 
 applied to transform the data and quantize the transformed 
 coefficients, balancing compression rate and quality. The 
 adaptive bitrate control strategy uses the 
 compression level and data queue length as input, 
 adjusting the compression level for the next point cloud 
 frame. Decoding is the inverse process of 
 compression, allowing the original point cloud to be 
 reconstructed using the corresponding inverse transforms.

\subsection{Point Cloud Encoding}
\subsubsection{Range Image Conversion} 
LiDAR sensors typically represent point clouds in Cartesian 
coordinates relative to the sensor origin, 
denoted as $(x,y,z)$. Spherical projection maps the point 
cloud from Cartesian coordinates into spherical 
coordinates, generating a 2D panoramic range image, 
denoted as $\mathcal{P}$. The formula for spherical 
projection from the original point cloud to a 2D range 
image is as follows:
\begin{equation}
\begin{aligned}
     i &= \left\lfloor {( \operatorname{atan2}(y, x)+h_{offset} )}/{\Delta\theta} \right\rfloor, \\
     j &= \left\lfloor ({\operatorname{atan2}(z,{\sqrt{x^2 + y^2 }})}+v_{offset})/{\Delta\varphi} \right\rfloor, \\
     r &= \sqrt{x^2 + y^2 + z^2},
\end{aligned}
\end{equation}
where indices $i$ and $j$ represent the pixel coordinates 
in the range image, and $r$ is the radial distance of 
the point cloud, which corresponds to the range value 
at $(i,j)$. $h_{offset}$ and $v_{offset}$
 are horizontal and vertical offsets, respectively, 
 related to the LiDAR's field of view (FOV), ensuring 
 that $i,j$ are non-negative. $\Delta\theta$ and 
 $\Delta\varphi$ represent the discretization granularity, 
 depending on the horizontal and vertical resolution of the 
 selected LiDAR.$\left\lfloor \cdot\right\rfloor$ represents the floor function.
\par
The range image is a compact representation 
that allows the simplification of 3D point cloud operations 
into 2D image operations, improving computational efficiency. 
Additionally, neighboring points in the range image are more 
likely to belong to the same range surface, 
enabling better exploitation of spatial relationships 
for encoding.

\subsubsection{Surface Encoding}
In practice, point cloud compression will 
choose an appropriate spatial modeling method 
(e.g., using planes or points\cite{c5}) to 
extract the structural information of point clusters,
reducing spatial redundancy. 
In fact, many points in real-world point clouds 
are located on the same plane (e.g., ground or walls), 
allowing these point clouds to be approximated by planes.
A point cloud in the range image can be represented as $p_{i,j,r}=(i,j,r)$, and its Cartesian coordinates can be written as:
\begin{equation}
    p_{x,y,z} = 
    ({r \cdot \cos\varphi \cdot \cos\theta
    , r \cdot \cos\varphi \cdot \sin\theta , r \cdot \sin\varphi}),
\end{equation}
where $\theta=i \cdot \Delta\varphi - h_{offset}$, and $\varphi=j \cdot \Delta\varphi - v_{offset}$ are the azimuth and elevation angles in the spherical coordinate system derived from the range image. 
The plane equation in Cartesian coordinates is given 
by $a \cdot x + b \cdot y + c \cdot z + d = 0$. 
We use the least squares method\cite{orlandini2002fitting} to fit the plane on 
which the point cloud lies. Then, the plane model can be used 
to predict the radial range $r$ of the point cloud. 
Given $\theta$ and $\varphi$, the predicted $r$ value, 
denoted as $\hat{r}$, can be obtained using the following 
equation:
\begin{equation}
\hat{r} = -\frac{d}{a \cdot \cos\varphi \cdot \cos\theta + b \cdot \cos\varphi \cdot \sin\theta + c \cdot \sin\varphi}.
 \label{eq:predict_r}
\end{equation}

Previous works \cite{c5}\cite{spatio-temporal} have employed such plane models, 
but the plane model is not the optimal choice since it cannot
 directly use $(i,j)$ to predict $\hat{r}$ without expensive 
 trigonometric computations. A more intuitive method is to 
 use Eq. \ref{eq:surface_model}, which we refer to as the surface 
 model. Though the surface model is not a plane in Euclidean 
 space, it better captures the spatial structure of points in 
 the range image, and it is easier to compute:
\begin{equation}
    \hat{r} = -\frac{d}{a \cdot i + b \cdot j + c }.
    \label{eq:surface_model}
\end{equation}

\begin{figure}
    \centering
    \includegraphics[width=1.0\linewidth]{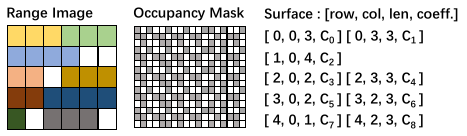}
    \caption{The range image is divided into macroblocks,
     with different colors representing 
     different surfaces (left). The occupancy 
     mask marks the location of the point cloud 
     in the range image, and the surface block is 
     encoded using a four-tuple (right).}
    \label{fig:plane_fit}
\end{figure}

Inspired by this, as shown on the left side of Fig.~\ref{fig:plane_fit}, we divide the range image into
  macroblocks (e.g., $4 \times 4$) 
  and iteratively fit the surface for each block. 
  We set a distance threshold $\Delta r$. 
  Only when all points in the block have distances 
  to the surface less than $\Delta r$ is the block 
  considered a surface. For blocks in the same row, 
  we use the surface parameters of the previous block to 
  predict the next block and perform the $\Delta r$ test.
   If the test passes, the blocks are merged and 
   share the same surface parameters. To reconstruct the range 
   image, as shown on the right side of Fig. \ref{fig:plane_fit}, 
   we record the positions of the point clouds 
   in the range image as an occupancy mask, and we encode 
   the surface using a four-tuple $(row, col, len, coefficients)$, 
   recording the position and parameters of the surface block.

\subsubsection{Unfit Points Encoding}
\par
After surface encoding, there will inevitably be some point clouds that are not fitted. 
The range image without all fitted surface points is called the 
unfit image. 
We can directly encode the original range values of unfit image,
as in\cite{spatio-temporal}. 
However, this will cause redundancy in accuracy because the surface 
model fitting has already relaxed the level of accuracy. 
Alternatively, it's a choice to use image encoding techniques (e.g., JPEG), 
but encoding the entire unfit image is not efficient as it 
contains a large number of zero-value pixels, which are 
redundant. Moreover, image compression may introduce 
zero-value noise \cite{chenghui50}.
\par
To only encode the areas of interest 
without introducing zero-value noise, the Shape-Adaptive DCT (SA-DCT) 
is applied on the unfit image to perform 
temporal and frequency domain transformations.
The core concept of SA-DCT for compressing 
the unfit point cloud is shown in Fig.~\ref{fig:banner}.
During encoding, 
only the remaining unfit points in the unfit image are 
shifted to the upper edge of the column and 
undergo a 1D DCT transformation.  
The 1D DCT transformation on the column(row)
vector $\mathbf{x}$ produces 
the 1D vector $\mathbf{c}_k$ as follows:
\begin{equation}
  \textbf{c}_k=A_{L_k} \cdot \mathcal{DCT}_{L_k} \cdot \textbf{x}_k.
\end{equation}

After that, the transformed non-zero elements are shifted to the 
left edge by rows, and a one-dimensional DCT transformation is 
performed.
Finally, we obtain a coefficient matrix $\mathcal{C}$.
The corresponding inverse transformation formula is as follows:
\begin{equation}
    \textbf{x}_k=\frac{2}{A_{L_k} L_k} \cdot \mathcal{DCT}_{L_k}^T \cdot c_k,
\end{equation}
where $L_k$ is the length of the vector $\mathbf{x}$, $A_{L_k}$ is the normalization factor, and the definition of $\mathcal{DCT}_{L_k}$ is as follows:
\begin{equation}
\begin{split}
    &\mathcal{DCT}_L(p, k) = a_0 \cdot \cos ( p ( k + \frac{1}{2} ) \frac{\pi}{L_k} ),\\
    &a_0 = \begin{cases} 
        \sqrt{\frac{1}{2}}, & \text{if } p = 0 \\
        1, & \text{otherwise.}
        \end{cases}
\end{split}
\end{equation}

After transformation, we quantize the coefficient matrix 
$\mathcal{C}$ by the quantization step $q_{step}$ to obtain
 the quantized coefficient matrix $\mathcal{C}^*$ (* denotes the quantized result). 
 Quantization introduces quantization errors but 
 reduces the data size, increasing the compression 
 rate of the entropy encoder\cite{collet2016zstandard}.
 During SA-IDCT reconstruction, 
 the position of non-zero elements in the original data 
 is needed for the inverse transform. 
 However, no additional storage is required because 
 the non-zero elements' positions can be derived 
 from the occupancy mask by excluding the fitted points' 
 mask.

\subsection{QoE-based Adaptive Bitrate Control}
In video streaming, 
ABR allows the video stream 
to dynamically adjust its quality based on network 
conditions, thus avoiding stuttering due to network 
fluctuations and improving the QoE\cite{c16}. 
For point cloud transmission used in cloud services, 
the server aims to receive small, 
high-quality point clouds from robots in real-time. 
We define a QoE objective function related to real-time 
transmission quality for cloud services, 
formulated as follows:
\begin{equation} \label{eq:qoe}
    QoE= \sum_{i=1}^{n} q(R_i) - \mu \sum_{i=1}^{n} K_i 
    - \sum_{i=1}^{n-1} \left| q(R_{i+1}) - q(R_i) \right|
    \end{equation}
in which the first term $q(\cdot)$ assigns 
a quality score based on compression level, and 
$R_i$ represents the configuration  
parameter of the compression level used for the i-th frame point cloud compression,
including vertical and horizontal resolution, surface threshold, quantization level.
The second term serves as a penalty for the length of the buffer queue, 
where $K_i$ represents the length of the buffer queue of the data packet 
encoded by the i-th frame point cloud waiting to be sent; 
$\mu$ is the corresponding weight. 
The third term is a penalty for switching quality, 
which is utilized to avoid frequent quality switching.

By optimizing the \eqref{eq:qoe}, the compression level configuration throughout the entire transmission process
 $R_{i:N}^{*}$ can be sovled as: 
\begin{equation*}
\begin{split}
&R_{i:N}^{*}=\arg\min_{R_{i:N}}~QoE\\
&s.t.\quad  R_i \in \left\{C_1, C_2, \ldots \right\}.
\end{split}
\end{equation*}
To address this online optimization problem, we propose a strategy based on the following schemes:
\begin{itemize}
    \item Quality improvement attempts: If the queue remains stable over a long period, we attempt to improve the quality (the first term of QoE). If the result is not satisfactory, we revert to the previous state.
    \item Historical memory: Decisions are made not only based on the current state but also by considering the historical buffer length (related to the second term).
    \item Buffer switching: Quality switching does not immediately affect the buffer length. Hence, a buffer period is necessary between switches to reduce the penalty from frequent quality changes (reduce the third term).
\end{itemize}
These schemes form the basis of our QoE-based adaptive bitrate control strategy, which optimizes transmission quality while minimizing latency.
\section{EXPERIMENTS}

\subsection{Experimental Setup}
The baseline point cloud compression methods selected for comparison are:
the KDTree-based method from Google: Draco \cite{c4};
Geometry-based compression method: G-PCC \cite{c10}, \cite{c11};
Range image-based compression using JPEG2000 (JPEG Range)\cite{christopoulos2000overview};
the octree-based method from the Point Cloud Library (PCL)\cite{PCL}.

Our experiments were conducted on a desktop platform 
with an Intel i7-12650H processor and 16 GB of RAM. 
For localization\cite{liu2023translo}, we used the LiDAR odometry  
KISS-ICP\cite{kissicp} on the KITTI odometry dataset\cite{kitti}. For object detection, 
we used PointPillar\cite{pointpillar} on the KITTI detection dataset.
For mesh reconstruction, we used VDBFusion\cite{vdbfusion} to extract meshes 
via TSDF maps and marching cubes on the MaiCity dataset. 
To comprehensively evaluate performance, we tested all methods 
across all sequences and frames and calculated the arithmetic mean.
Our method's parameters include 
($\Delta \theta$, $\Delta \varphi$, $\Delta r$, $q_{step}$), 
and in subsequent experiments, 
we use a four-tuple to represent the parameter settings.

For QoE-based Adaptive Bitrate Control, 
we manually simulated network bandwidth variations 
using a local area network (LAN) router and recorded 
the compression quality and buffer queue length at each time step, 
computing the QoE for the entire transmission process.
\subsection{Comparative Results}
\par\textbf{Mesh Reconstruction}
Our compression method outperformed other methods, 
achieving better mapping performance and higher compression ratios.
we use the F-score to evaluate surface 
quality in mesh reconstruction and the compression ratio is computed as 
original point cloud size divided by the final binary size.
\par
\begin{figure}[!htbp]
    \centering
    \includegraphics[width=1\linewidth]{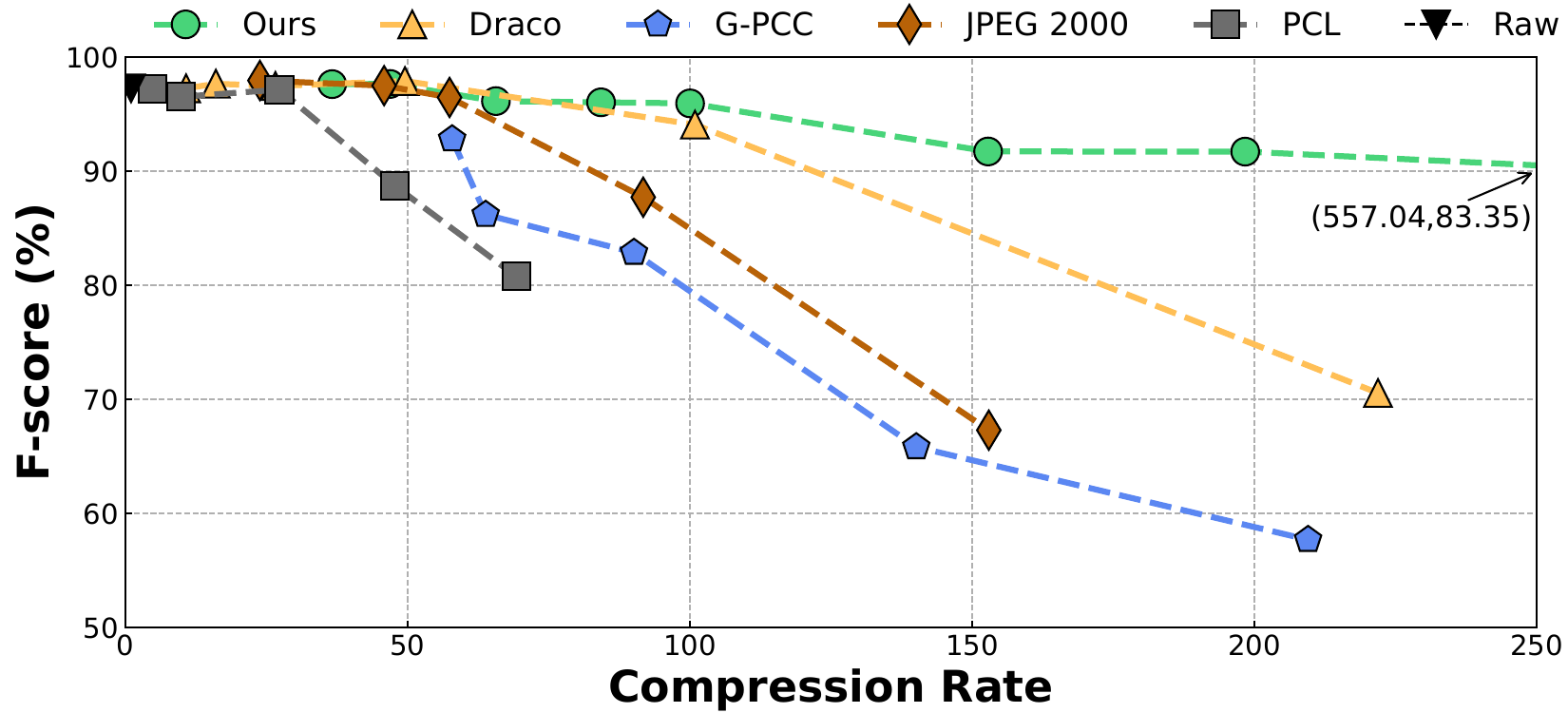}
    \caption{The F-score of mesh reconstruction and compression rate comparison of various methods.}
    \label{fig:fscore}
\end{figure}

Fig. \ref{fig:fscore} compares the F-scores of different methods at various compression ratios.
Our method can achieve a compression ratio of 100.5$\times$ and maintain an F-score of 95.94\%.
At a compression ratio of 198.4$\times$, our method's F-score remains at 91.70\%.
In comparison, the best-performing Draco achieved an F-score of 94.07\% at a compression ratio of 100.7$\times$,
and as the compression ratio further increases, the performance of other methods' F-scores drops rapidly,
while our method still maintains a stable F-score performance. Our method effectively 
combines and balances quantization and down-sampling strategies, even at a very high compression ratio 
(even at 557$\times$), it can still maintain a stable F-score.
\par

\begin{figure}[!htbp]
    \centering
    \includegraphics[width=1\linewidth]{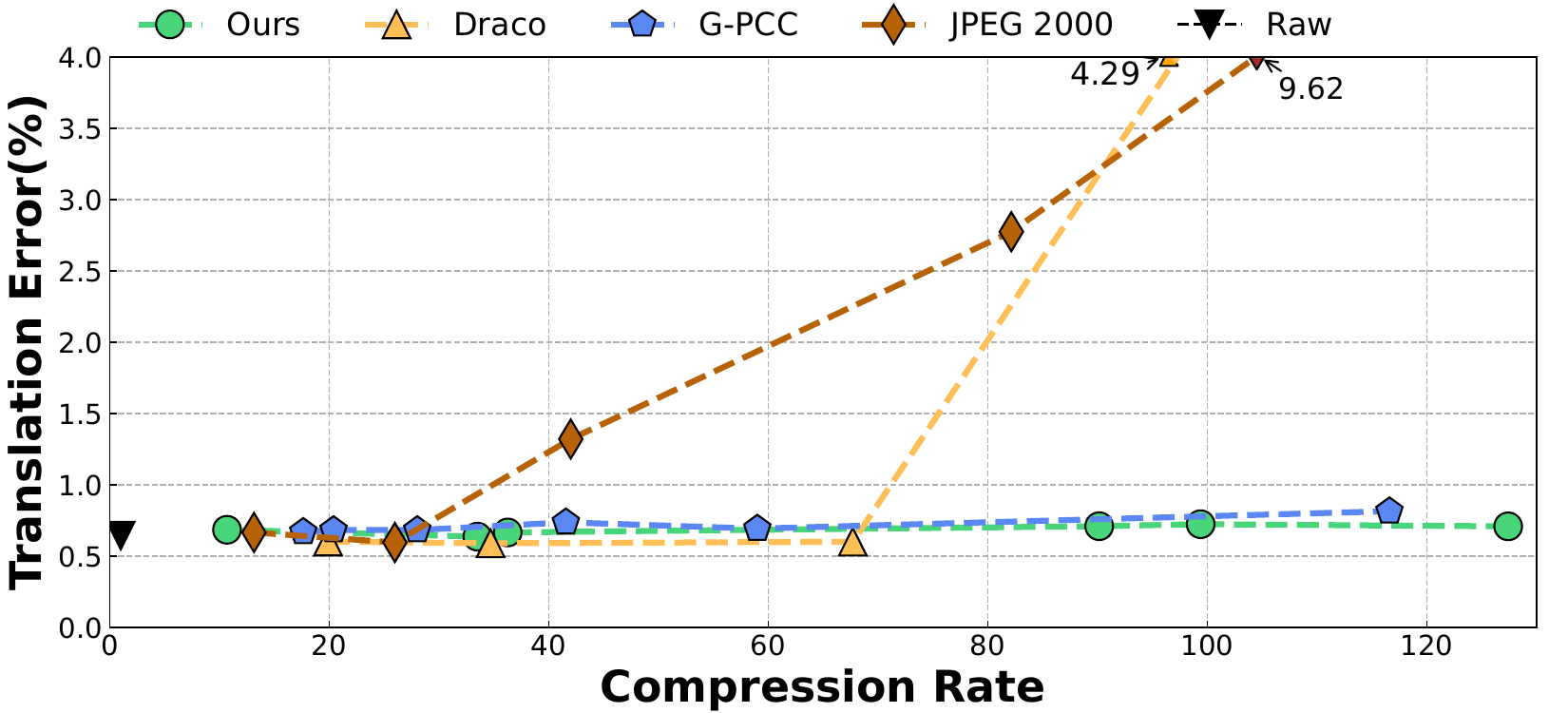}
    \caption{Average translation error and compression rate  comparison of various methods.}
    \label{fig:translation}
\end{figure}

\textbf{Localization}
Our compression method has an average translation error($\%$)\cite{kissicp} 
lower than other baseline methods when the compression 
ratio is greater than 100$\times$. Fig. \ref{fig:translation} shows the Translation
Error($\%$) of different methods at different compression
ratios on the KITTI dataset (seq.00\textasciitilde10). We did not show 
the results of the PCL because the PCL failed in this test.
In Fig. \ref{fig:translation}, 
our method achieves 36.3$\times$ compression rate 
while maintaining almost the same Translation Error (\%) 
as the original point cloud. As the compression rate increases, 
our method's localization performance remains stable, 
while JPEG2000 and Draco show a significant increase 
in Translation Error (\%) to 4.29\% and 9.62\% at around 100$\times$. 
In contrast, our method maintains a low error rate 
of 0.71\% at a 99.7$\times$ compression rate, 
outperforming all other baseline methods.
\par

\begin{figure}[!htbp]
    \centering
    \includegraphics[width=1\linewidth]{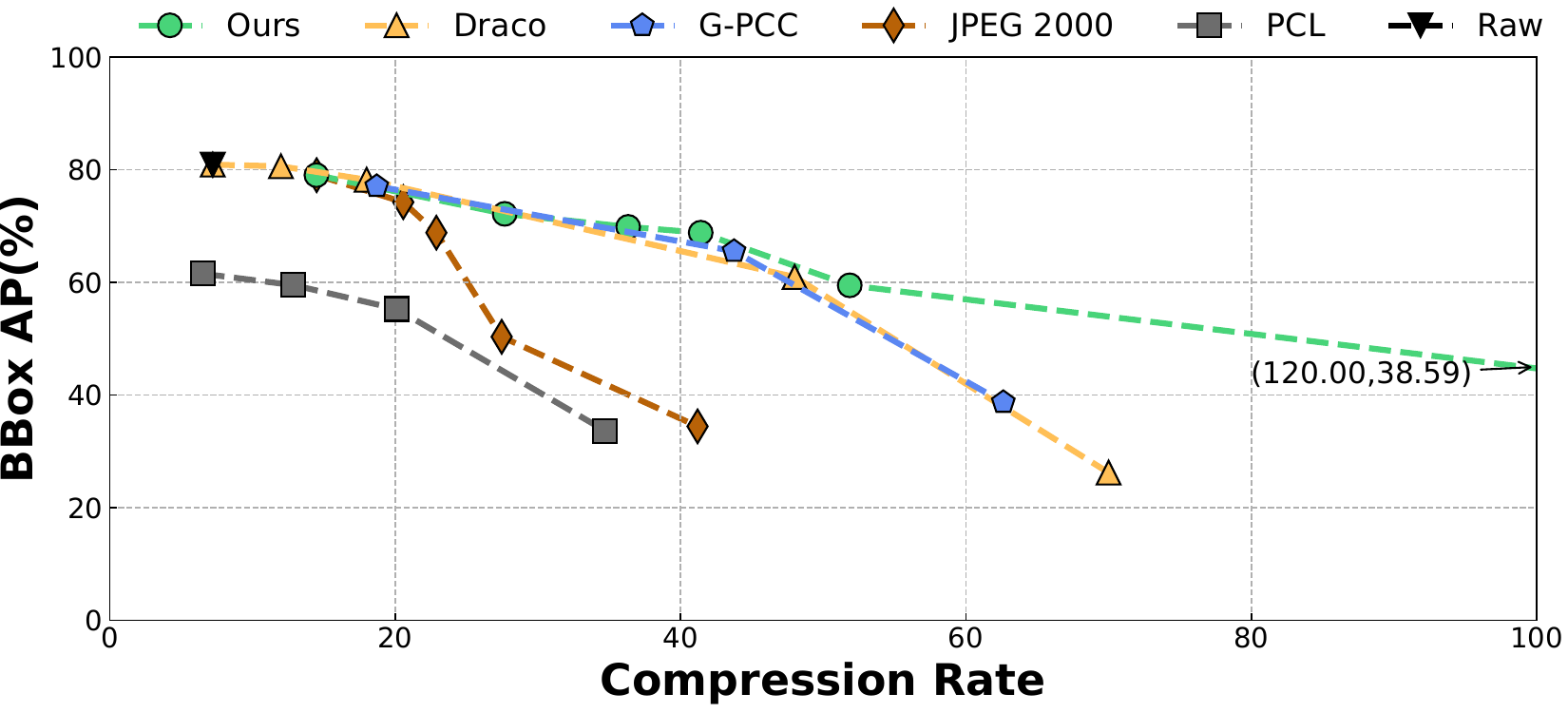}
    \caption{The object detection BBox AP and compression rate comparison of various methods.}
    \label{fig:bbox}
\end{figure}

\textbf{Object Detection}
Our method achieves higher object detection 
accuracy than the other methods when the 
compression rate exceeds 36.4$\times$.The overall bounding box average precision (AP) 
for annotated classes\cite{pointpillar} is evaluated.
Fig. \ref{fig:bbox} shows 
BBox AP(\%) on the KITTI object detection 
dataset for different methods at various 
compression rates.Compared to the original point cloud's 80.89\% 
accuracy, our method achieves a competitive 
79.03\% accuracy at a compression rate of 14.4$\times$. 
As the compression rate increases, our method 
maintains 68.8\% accuracy at a compression rate 
of 41.4$\times$. In contrast, PCL and JPEG2000's accuracy 
drops rapidly, falling below 40\% at compression 
rates greater than 40$\times$. 

\subsection{Ablation Study}
To verify the impact of the modules 
mentioned earlier, we conducted experiments on 
1,000 frames from the KITTI dataset 
and recorded the relevant data. 
All experiments were conducted with 
parameter settings \textbf{(0.5, 0.5, $\cdot$ , $\cdot$)}.

\par
\textbf{Plane Model and Surface Model.}
Tab.\ref{tab:model_comparation} shows the mean average error (MAE) of 
point clouds predicted by plane and surface models compared to the ground 
truth in the range image at various distance thresholds. Under the same 
distance threshold, surface models produce lower MAE than 
plane models.
\par
\textbf{SA-DCT}
Tab.\ref{tab:ablation} compares the MAE and corresponding compression rate between no compression of the unfit image (No Compr.) and compressing the unfit image using SA-DCT.  
With No Compr. (generated only by surface model fitting),the pipeline achieves 32.40$\times$ compression rate with an MAE of 3.68.  
With SA-DCT, the compression rate improves to 40.86$\times$ when the quantization step is 0.10, with a slight increase in MAE to 5.29.  
The experiments demonstrate that our method achieves a significant improvement in compression rate with an acceptable increase in error.
\begin{table}[htbp]
	\scriptsize
	\setlength\tabcolsep{7pt}
	\centering	
    \caption{Fitted points MAE comparison between Plane Model and Surface Model, MAE unit: $cm$.}
    \label{tab:model_comparation}
    \begin{tabular}{lcccccc}
            \toprule
    \multicolumn{1}{c}{Model}&\multicolumn{3}{c}{Plane } & \multicolumn{3}{c}{Surface} \\
            % \specialrule{1pt}{1pt}{3pt}
            \midrule
            Threshold&0.1m &0.3m &0.5m & 0.1m &0.3m &0.5m \\
            Fitted MAE &5.97&13.46&16.70&4.75& 10.82&11.92\\
            \bottomrule
    \end{tabular}
\end{table}

\begin{table}[htbp]
	\scriptsize
	\setlength\tabcolsep{8pt}
	\centering	
    \caption{Comparison results of using SA-DCT on unfit points and no compression. CR: Compression Rate, QS: Quantization Step, MAE Unit: $cm$.}
    \label{tab:ablation}
\begin{tabular}{lccc}
\toprule
    Configuration & Unfit MAE & MAE & CR \\
\midrule
No Compr     & 0.19      & 3.68  &  32.40  \\
SA-DCT(0.10) & 4.23   & 5.29  & 40.86   \\
SA-DCT(0.40) & 16.71  & 10.30 & 47.19  \\
SA-DCT(1.00) & 41.28  & 20.05 & 53.37   \\
\bottomrule
\end{tabular}
\end{table}
\subsection{Transmission Experiments}
\begin{figure}
    \centering
    \includegraphics[width=1.0\linewidth]{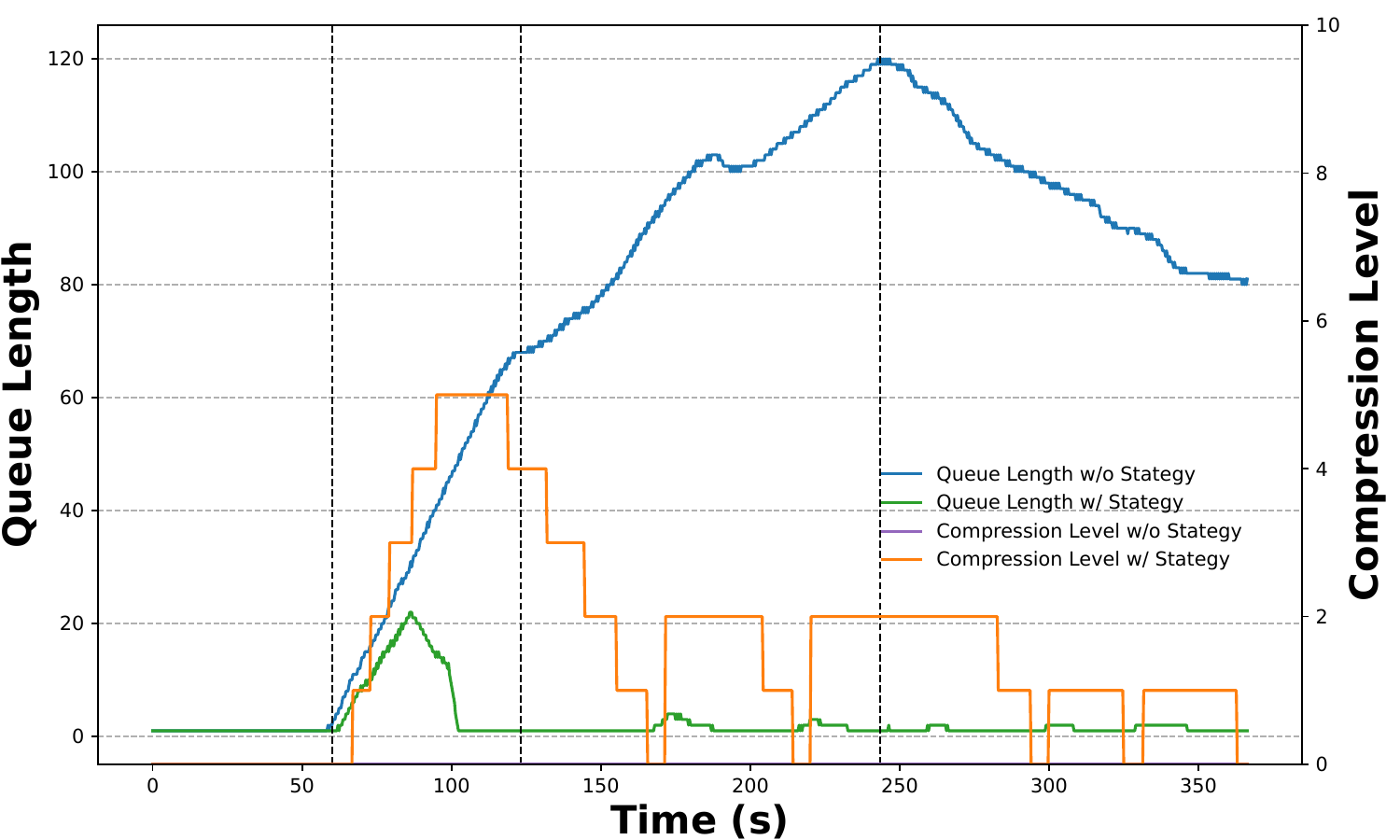}
    \caption{Comparison of the experiment results for point cloud transmission with and without the QoE-based adaptive bitrate control strategy.
    }
    \label{fig:strategy}
\end{figure}

For Transmission, 
we evaluate the QoE during point cloud transmission, which is critical for the quality of cloud services.
We compare the transmission performance of our point cloud compression method with and without the strategy.
where the compression level is predefined from fine 
to coarse (0 to 5) 
based on different parameters. 
In Fig.~\ref{fig:strategy}, the time points where the bandwidth changes are 
marked with vertical dashed lines. 
The experiment starts with a bandwidth of 300KB/s, 
which drops sharply to 100KB/s at 55 seconds 
(simulating interference or the robot entering 
a closed building where signal quality degrades); 
at 120 seconds, the bandwidth increases slightly to 130KB/s;
 and at 245 seconds, the bandwidth rises 
 to 160KB/s (indicating that the robot has moved away 
 from interference or exiting the closed building).  
\par  
In Fig.~\ref{fig:strategy}, when the bandwidth decreases due to the buffer switching characteristic of 
our adaptive bitrate control strategy, 
the compression level gradually increases 
from 0 to 5 to adapt to the bandwidth change. 
As a comparison, the compression level of without strategy 
remains at 0 during bandwidth drops, causing the data queue to grow and 
eventually leading to increased transmission delays. 
When the bandwidth rises at 125 seconds, 
our strategy gradually lowers the compression 
level (through the quality improvement attempts) 
to provide higher-quality point clouds. 
However, this quality improvement can cause delays, 
and at 160 seconds, reducing the compression level 
to 0 results in an increased queue length. Our 
historical memory scheme prevents the overuse of 
the quality improvement attempts by 
rolling back the compression level, 
maintaining stability in the data queue.  
\par  
To calculate QoE, 
we set the quality evaluation function 
as $q(i)=25-5i, i\in\left\{0...5\right\}$ 
and $\mu=0.5$. Throughout the point cloud 
transmission, the average QoE of the method 
with strategy is 18.33, while without strategy, the average QoE is 60.18. 
Our method maintains a shorter data queue 
length throughout transmission, 
reducing transmission delay.
\begin{table}[]  
    \centering  
    \caption{Average time consumption (ms) of different methods on the kitti dataset}  
    \label{tab:time_comparison}  
    \begin{tabular}{lccc}  
        \toprule  
        Method & Encode (ms) & Decode (ms) & Total (ms) \\  
        \midrule  
        G-PCC & 445.41 & 238.93 & 684.34 \\  
        Draco  & 49.08 & 10.68 & 59.76 \\  
        JPEG  & 36.37 & 13.79 & 50.13 \\
        PCL  & 144.90 & 86.43 & 231.33 \\
        \textbf{Ours}   & 41.06 & 11.35 & 52.41 \\
        \bottomrule  
    \end{tabular}  
\end{table}  
\begin{figure}[!ht]
    \centering
    \includegraphics[width=0.85\linewidth]{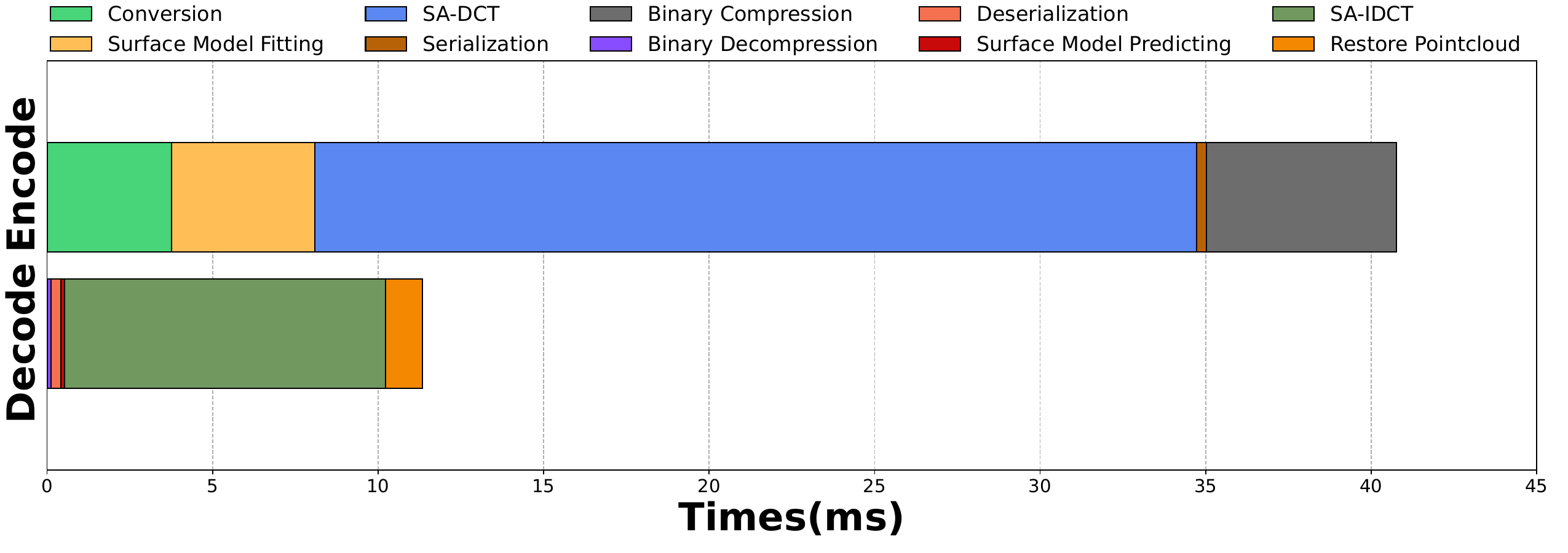}
    \caption{The module runtime breakdown of our method.
    }
    \label{fig:run_time}
\end{figure}

\subsection{Runtime Analysis}
Tab.~\ref{tab:time_comparison} compares the runtime of our method against other 
baseline methods. Our method's parameter settings are (0.5, 0.5, 0.3, 0.2), 
and all methods were set to achieve similar compression rates 
(approximately 60$\times$). Our method's encoding time is 41.05 ms, 
and decoding time is 11.35 ms. The speed of our method is sufficient 
to support real-time point cloud compression for current LiDAR.

Fig.~\ref{fig:run_time} provides a detailed runtime breakdown 
of our method's encoding and decoding processes. 
The most time-consuming components of the pipeline are 
SA-DCT and its inverse transformation.

\section{CONCLUSIONS}

This paper presents a novel real-time LiDAR 
point cloud compression and transmission framework, called \textit{RCPCC}. 
The combination of efficient point cloud compression 
and adaptive bitrate control strategy enables 
our method to help resource-constrained robots 
achieve real-time point cloud transmission,  
Our method achieves a 
compression rate of up to 80x, with real-time compression speed ($>$ 10 FPS) while maintaining high application accuracy. It surpasses state-of-the-art point cloud compression standards in terms of compression rate, speed, and accuracy.

   % This command serves to balance the column lengths
                                  % on the last page of the document manually. It shortens
                                  % the textheight of the last page by a suitable amount.
                                  % This command does not take effect until the next page
                                  % so it should come on the page before the last. Make
                                  % sure that you do not shorten the textheight too much.

%%%%%%%%%%%%%%%%%%%%%%%%%%%%%%%%%%%%%%%%%%%%%%%%%%%%%%%%%%%%%%%%%%%%%%%%%%%%%%%%

%%%%%%%%%%%%%%%%%%%%%%%%%%%%%%%%%%%%%%%%%%%%%%%%%%%%%%%%%%%%%%%%%%%%%%%%%%%%%%%%

%%%%%%%%%%%%%%%%%%%%%%%%%%%%%%%%%%%%%%%%%%%%%%%%%%%%%%%%%%%%%%%%%%%%%%%%%%%%%%%%

\bibliographystyle{IEEEtran} 
\bibliography{arxiv}

\end{document}